\documentclass[preprint]{elsarticle}
\makeatletter
\def\ps@pprintTitle{%
	\let\@oddhead\@empty
	\let\@evenhead\@empty
	\def\@oddfoot{\footnotesize\centerline{\thepage}\hfil
		\today}
	\let\@evenfoot\@oddfoot
}
\makeatother

\usepackage{url}
\usepackage{tabularx}
\usepackage{changepage}
\usepackage{algorithm}
\usepackage{algpseudocode}
\usepackage{algorithmicx}
\usepackage{multirow}
\usepackage{rotating}
\usepackage{booktabs}
\usepackage{mwe}
\usepackage{subfig}
\usepackage{xcolor}
\usepackage[utf8]{inputenc} 
\usepackage[T1]{fontenc}    
\usepackage[hidelinks]{hyperref}       
\usepackage{url}            
\usepackage{booktabs}       
\usepackage{amsfonts}       
\usepackage{nicefrac}       
\usepackage{microtype}      
\usepackage{lipsum}
\usepackage{graphicx}
\graphicspath{ {./images/} }

\usepackage{url}
\usepackage{tabularx}
\usepackage{changepage}
\usepackage{algorithm}
\usepackage{algpseudocode}
\usepackage{algorithmicx}
\usepackage{multirow}
\usepackage{rotating}
\usepackage{booktabs}
\usepackage{mwe}
\usepackage{xcolor}
\usepackage{amsmath}
\usepackage{amssymb}
\usepackage{textcomp}

\algnewcommand{\Input}[1]{%
	\Statex \textbf{Input:}
	\Statex \hspace*{\algorithmicindent}\parbox[t]{.8\linewidth}{\raggedright #1}
}

\algnewcommand{\Output}[1]{%
	\Statex \textbf{Output:}
	\Statex \hspace*{\algorithmicindent}\parbox[t]{.8\linewidth}{\raggedright #1}
}

\makeatletter
\newcommand{\multiline}[1]{%
	\begin{tabularx}{\dimexpr\linewidth-\ALG@thistlm}[t]{@{}X@{}}
		#1
	\end{tabularx}
}
\makeatother

\usepackage{subfig}

\begin{document}
		\begin{frontmatter}
		\title{Complexity Measures and Features for Times Series classification}
		\author[DECSAI]{Francisco J.~Bald\'an\corref{cor1}}
		\ead{fjbaldan@decsai.ugr.es}
		\author[DECSAI]{Jos\'e M.~Ben\'itez}
		\ead{J.M.Benitez@decsai.ugr.es}
		
		\cortext[cor1]{Corresponding author}
		\address[DECSAI]{Department of Computer Science and Artificial Intelligence, University of Granada, DICITS, iMUDS, DaSCI, 18071 Granada, Spain}

		\begin{abstract}
		Classification of time series is a growing problem in different disciplines due to the progressive digitalization of the world. Currently, the state-of-the-art in time series classification is dominated by The Hierarchical Vote Collective of Transformation-based Ensembles. This algorithm is composed of several classifiers of different domains distributed in five large modules. The combination of the results obtained by each module weighed based on an internal evaluation process allows this algorithm to obtain the best results in state-of-the-art. One Nearest Neighbour with Dynamic Time Warping remains the base classifier in any time series classification problem for its simplicity and good results. Despite their performance, they share a weakness, which is that they are not interpretable. In the field of time series classification, there is a tradeoff between accuracy and interpretability. In this work, we propose a set of characteristics capable of extracting information on the structure of the time series to face time series classification problems. The use of these characteristics allows the use of traditional classification algorithms in time series problems. The experimental results of our proposal show no statistically significant differences from the second and third best models of the state-of-the-art.  Apart from competitive results in accuracy, our proposal is able to offer interpretable results based on the set of characteristics proposed.
			
		\end{abstract}
		
	
 		\end{frontmatter}
	
\section{Introduction}
\label{sec:introduction}
At present, large amounts of information are recorded from a wide variety of fields. There is a growing need to analyze and classify these data to obtain useful information, for example, to identify different patterns of electricity consumption in order to adapt prices to consumers~\cite{markovivc2019data}, to identify cardiac anomalies characteristics of a pathology~\cite{chauhan2019ecg} or search for anomalies in starlight curves~\cite{twomey2019application}.

The field of time series classification (TSC)~\cite{bagnall2017great} has historically been dominated by proposals that offer good classification results but are hardly interpretable. For example, a simple approach that achieves good average results in the different types of problems is One Nearest Neighbour with Dynamic Time Warping (1NN+DTW)~\cite{berndt1994using}\cite{ratanamahatana2004making}.  This approach tells us how similar the time series are to each other, but it does not allow us to extract additional information from the problem. Recently the Collective Of Transformation Ensembles (COTE)~\cite{bagnall2015time} has been shown to obtain the best TSC results on the reference time series database collected in the UCR repository~\cite{UCRArchive}, in the 2015 version of this repository. This algorithm is composed of 35 classifiers (flat-COTE) which apply cross-validation on the training set. COTE contains reference classifiers in the fields of TSC. These classifiers are evaluated internally with cross-validation, and depending on their results, they are included in the final result. Recently The Hierarchical Vote Collective of Transformation-based Ensembles (HIVE-COTE)~\cite{bagnall2020usage} has been proposed, which improves the classification process carried out by its previous versions. HIVE-COTE is composed of several classifiers of different domains distributed in five large modules. Each module provides a probability estimate for each class and obtains a weighting proportional to the accuracy achieved over the training set. HIVE-COTE combines these estimates in a second layer and obtains the predicted class from the highest weight over all the modules. The HIVE-COTE proposal provides the best results, but its interpretability is very low, and its high computational cost prevents its application in large datasets.

Other more interpretable approaches as decision trees do not usually obtain competitive results in the field of TSC. This behavior is due to their inability to capture the time relationships between the different time instants that make up a time series. These approaches are successfully used in combination with other proposals, such as shapelets, which extract behavioral patterns from time series~\cite{ye2009time}. These patterns make it possible to differentiate time series belonging to different classes. These proposals have great interpretability since they allow us to identify, in a graphical way, patterns of interest belonging to the different classes that compose the problem. Although, in this case, there are also proposals such as the Shapelet Transform (ST)~\cite{lines2012shapelet}, which transforms these shapelets into features. ST alter the problem of TSC into a traditional, vector-based, classification problem, on which we can apply traditional algorithms, such as Random Forest (RF)~\cite{Breiman2001}, and obtain good results. In this way, there are proposals in Big Data such as Distributed FastShapelet Transform~\cite{baldan2018distributed} that allows us to face TSC problems in massive data environments where traditional TSC algorithms cannot be applied due to their high computational complexity. There is a more recent proposal, which proposes the creation of a weighted ensemble of standard classifiers, such as Random Forest, Naive Bayes, Support Vector Machines, among others, on the transformed data, obtaining very competitive results. This proposal is named Shapelet Transform Classifier (STC)~\cite{bostrom2017binary}.

In the literature, we can find proposals focused on extracting a large number of characteristics from time series~\cite{fulcher2013highly}\cite{fulcher2018feature}. The main idea of these characteristics is any type of mathematical operation applicable over a time series that provide valuable information. The objective of these proposals is to look for an underlying structure that represents the behavior of a time series. These types of studies are ambitious but difficult to interpret over a specific problem due to the high number of characteristics present. Moreover, these studies are oriented to the unsupervised learning environment. Some proposals make a selection of the main characteristics of a time series, from the theoretical point of view that could explain the origin of their behavior~\cite{kang2018efficient}. The objective of the previous work is to generate synthetics time series that represent real problem behaviors, so its main target is far from the problem of TSC. On the other hand, CAnonical Time-series CHaracteristics (catch22)~\cite{Henning2019catch22} proposes a set of 22 characteristics that have been selected based on the classification results obtained on a large set of datasets. For this proposal, a large number of characteristics and their possible combinations have been tested, measuring the classification results obtained. The main criterion for selecting the characteristics is to provide the best possible results, although the execution time and, in some cases, their interpretability is also taken into account. Recently a method has been proposed in this line. This method, called Feature and Representation Selection (FEARS)~\cite{bondu2019fears}, is based on obtaining different alternative representations such as derivatives, cumulative integrals, power spectrum, among others, of the time series. This method then extracts characteristics of interest using an automatic variable construction technique. As the last step, a Naive Bayes classifier is in charge of learning about the new extracted characteristics. This procedure is repeated several times to obtain the most informative set of characteristics possible.

There are other proposals based on studying the complexity of time series~\cite{zhou2017measuring}\cite{parmezan2015study}. These proposals use complexity measures that measure the interrelationships between the different values in a time series. A greater number of relationships lead to greater complexity. In the same way that the traditional characteristics of the time series are capable of providing sufficient information about a problem, complexity measures can add useful information to the problem.

In this work, we present a set of characteristics composed of complexity measures and representative features of the time series. This transformation allows the use of traditional learning algorithms on TSC problems. Additionally, this transformation allows interpreting the results obtained by the classification algorithms. The performance of our proposal has been tested on a set of 112 datasets present in the UCR repository. We have applied the most popular and widely used classification algorithms based on trees that allow interpretable results. Our proposal is publicly available as an R package in the online repository~\footnote{Complexity Measures and Features for Times Series classification. \url{https://github.com/fjbaldan/CMFTS/}}

The rest of the work is organized as follows: Section~\ref{background} introduces the state-of-the-art in TSC: distance-based methods, feature-based methods, and deep learning methods. In Section~\ref{CMF_proposal}, we describe in depth our proposal. Section~\ref{Exp} shows the experimental design used, the results obtained, and the interpretability of these results. Finally, Section~\ref{Concl} concludes the paper.

\section{Related work}
\label{background}

There are several ways to group the TSC algorithms. In this work, we group them by the type of data on which each algorithm works and its internal operation. In this way, we have three principals groups with their corresponding subgroups. A first group is composed of the distance-based proposals (Section~\ref{distancebased}), which are strongly related to calculations of similarity and distance between different time series or subsequences of the time series themselves. A second group is composed of features-based proposals (Section~\ref{featurebased}), which are based on the calculation of certain parameters of the time series that transform the original data.  After this transformation, traditional classification algorithms are applied to the new dataset. The last group would be made up of the deep learning proposals (Section~\ref{deeplearning}), where data entry and processing depend entirely on each proposal.

\subsection{Distance-based Classification}
\label{distancebased}

Patterns searched for in TSC problems may have their origin in different domains. For this reason, there are different types of approaches depending on multiple factors. There are currently six main approaches for dealing with this kind of TSC problems~\cite{bagnall2017great}, grouped by the type of discriminatory features that the technique attempts to find:

\begin{itemize}
	\item Proposals that use all the values of the time series: are linked to the use of similarity measures and different types of distance. The reference algorithm of this group is 1NN+DTW, which is simple to apply but has high computational complexity. This algorithm is often used as a benchmark in TSC problems.
	
	\item Those using phase-dependent intervals: they use small subsets from each time series, rather than using the entire time series. Proposals like Time Series Forest (TSF)~\cite{deng2013time} have been proved that extracting characteristics such as mean, variance, or slope from random intervals, and use them as classifier features,  works particularly well. Characteristics such as Fourier, autocorrelation, and partial autocorrelation, which are more complex and related to the time series than those mentioned above, are used by more recent proposals such as contract Random Interval Spectral Ensemble (c-RISE)~\cite{flynn2019contract}, with very competitive results.
	
	\item The independent phases, based on shapelets: the shapelets based ones look for substrings of the time series that allow differentiating the time series belonging to each class. They are closely linked to the use of similarity and distance measurements. The first proposals generated simple classification trees capable of differentiating the belonging of a time series to one class or another according to the presence or not of a certain subsequence in it~\cite{ye2009time}\cite{rakthanmanon2013fast}\cite{mueen2011logical}. These approaches offered some interpretability to the results. Recent work on shapelets has shown that they are particularly useful when used as input features to a traditional classification algorithm~\cite{lines2012shapelet}\cite{bostrom2017binary}\cite{baldan2018distributed}, rather than as part of the classification tree itself.
	
	\item Based on dictionaries: in some cases, the presence of a certain pattern in a time series is not enough to identify whether it belongs to one class or another~\cite{lin2009finding}\cite{schafer2015boss}. There are problems in which the number of times the pattern appears in a time series is determinant to classify it correctly. The shapelets are not useful in these cases, and the use of algorithms based on dictionaries is mandatory. These algorithms count both the presence or absence of each subsequence in a time series. They create a classifier based on the histograms obtained from these dictionaries. The way of creating the dictionary is one of the main differences among the proposals of this type. For example, Bag of patterns (BOP)~\cite{lin2012rotation} creates the dictionary through the Symbolic aggregate ApproXimation (SAX)~\cite{lin2007experiencing} words extracted from each window. Symbolic Aggregate approXimation-Vector Space Model (SAXVSM)~\cite{senin2013sax} combines the SAX representation used in BOP with the vector space model commonly used in Information Retrieval and counts the appearance frequencies over the classes and not over the time series. Bag of SFA symbols (BOSS)~\cite{schafer2015boss} does not use Piecewise Aggregate Approximation (PAA)~\cite{keogh2001locally} in its SAX transformation but uses truncated Discrete Fourier Transform (DFT). Furthermore, it uses the so-called Multiple Coefficient Binning (MCB) technique to discretize the truncated time series, among other differences. Despite the good results, BOSS does not scale well, so it made a proposal called contracted BOSS (cBOSS)~\cite{large2019time}, which modified the way BOSS classifiers were chosen, indicating construction time limits per classifier and saving the advances during the construction process without significant accuracy changes. Word ExtrAction for time SEries cLassification (WEASEL)~\cite{schafer2017fast} is one of the latest proposals made. WEASEL has highly competitive results and differs from the rest by its ability to derive the characteristics obtained, achieving a new, much smaller, and more discriminating set of features.
	
	\item Based on models: this approach is mainly oriented to problems with long time series, but with different lengths~\cite{bagnall2014run}\cite{chen2013model}. These proposals usually fit a model to each time series and measure the similarity between the models. It is an approach that is not sufficiently widespread and is applied to particular problems.
	
	\item Combinations or ensembles: this approach works both in time series and traditional classification problems, using the results of different models to make a final decision. In the area of time series, HIVE-COTE~\cite{bagnall2020usage} is the best proposal to date. It uses models from different approaches and offers the best accuracy results. On the other hand, it is the approach with the highest computational complexity due to the high number of algorithms it uses and its corresponding computational complexities. Moreover, this large number of algorithms leads to low interpretability of results.
\end{itemize}

Each of these approaches adapts to different types of problems, but they all work on the original values of the time series.

\subsection{Feature-based Classification}
\label{featurebased}

The feature-based approach is focused on a transformation to the time series dataset, obtaining a new dataset composed of different features that explain the behavior of the original time series~\cite{fulcher2018feature}. The feature-based approach offers multiple advantages over the distance-based approach for dealing with time-series classification problems. This approach allows analysis of time series on different time domains and with different lengths, being more widely applicable because the stationarity properties of the series are not always required. In addition, this approach allows us to use the standard classification and clustering methods that have been developed for non-time series data. In this approach, we can found two different approximations:


\begin{itemize}
	\item The first one is based on the use of a reduced set of characteristics with a strong theoretical basis that is easily interpretable. In addition to applying traditional learning algorithms to the problem, this approach offers the possibility to analyze the extracted parameters and to obtain additional information.
	
	Based on this approach, we can find proposals that, with a minimum of four initial characteristics such as mean, typical deviation, skewness, and kurtosis, are able to face the problem of the classification synthetic control chart patterns used in the statistical process control~\cite{nanopoulos2001feature}. There are also proposals, focused on the improvement of accuracy, based on the creation of an ensemble for classification, composed of trained classifiers on different representations of the time series~\cite{bagnall2012transformation}: power spectrum, autocorrelation function, and a principal components space. The final classification is obtained from a weighted voting scheme. In the field of clustering, some proposals use characteristics of time series such as trend, seasonality, non-linearity, among others, which are very appropriate to express the behavior of a time series~\cite{wang2006characteristic}.
	
	In this approach, we also find proposals that aim to generate synthetic time series with a given behavior as close as possible to a real time series~\cite{kang2018efficient}. This work contains a selection of the main characteristics of a time series. Its objective is to use them to generate time series with a real behavior with these controllable parameters.
	
	\item The second approach focuses on applying a large number of different operations to obtain a great set of descriptive parameters of the time series analyzed. In this approach, the selection of the characteristics of interest resides in the learning algorithm used on the transformed dataset. Having a much greater set of characteristics than the first approach allows us to capture a higher number of behaviors of interest, improving the results of the algorithms applied afterward. But it is hard to extract useful information because there are a large number of characteristics to analyze. In addition, it is possible that a large part of the selected characteristics is not as explanatory as the characteristics with a strong theoretical base such as trend, seasonality, among others.
	
	In this area, we can find different proposals. For example, the use of 8,651 operations on a set of 875 time series~\cite{fulcher2013highly}, coming from different fields, with the aim of extracting the different possible structural behaviors. Another of its objectives is to find possible interrelations between time series coming from different fields. Given the rearrangement of the rows (original time series) in the final matrix of characteristics, based on the similarity between the different operations calculated, this work can be included within the field of clustering. Another objective of the previous work would be to find a shared underlying structure between time series belonging or not to the same scope.
	
\end{itemize}

In a more controlled environment, within the reference problems of classification of time series, we found a similar proposal to the previous one. In this case, the authors seek to obtain the best classification results by working on the transformed dataset~\cite{fulcher2014highly}. It has almost 9,000 characteristics, being of special importance the way to select the variables of interest. This proposal opted for the selection of the combination of variables that offers the best classification results, using the following procedure:

In the first place, the proposal selects the variable that obtains the best classification result by itself. Then, one by one, it combines the previously selected variable with the rest of the variables, and the variable that offers the best results is selected as the second variable. This set of two variables is then combined with each of the other variables and evaluated. This process is repeated until the stop criterion is met. However, this proposal entails a high computational complexity due to a large number of combinations available.

\subsection{Deep Learning Classification}
\label{deeplearning}
The approach based on deep learning has gained popularity recently~\cite{fawaz2019deep}. Although it is usually related to the processing of images, it has very interesting proposals in the field of TSC~\cite{wang2017time}. We can distinguish between two main groups inside this approach: Generative Models and Discriminative Models.

\begin{itemize}
	
	\item In the Generative Models, there is usually a previous step of unsupervised training to the learning phase of the classifier. Depending on the approach, two subgroups can be differentiated: Auto Encoders and Echo State Networks. In the case of Auto Encoders, there are a large number of proposals, for example, to model the time series before the classifier is included in an unsupervised pre-training phase such as Stacked Denoising Auto-Encoders (SDAEs)~\cite{bengio2013generalized}. A Recurrent Neural Network (RNN) Auto Encoder~\cite{rajan2018generative} was designed to generate time series first and then use the learned representation to train a traditional classifier. After that, it predicts the class of the new input time series. A model based on Convolutional Neural Networks (CNN)~\cite{song2020representation} was proposed where the authors introduced a deconvolutionary operation followed by an upsampling technique that helps to reconstruct a multivariate time series. In the case of the Echo State Networks, these networks were used to reconstruct time series and use the representation learned in the space reservoir for classification. They were also used to define a kernel on the learned representations and apply an MLP or SVM as a classifier.
	\item In the case of Discriminative Models, these are a classifier or regressor that learns the mapping between the input values of the time series and returns the probability distribution over the class variable of the problem. In this case, we can differentiate two subgroups: Feature Engineering and End-to-End. The typical case of use of Feature Engineering is the transformation of the time series into images, using different techniques such as recurrence plots~\cite{hatami2018classification} and Markov transition fields~\cite{wang2015spatially}, and introduce that information in a deep learning discriminating classifier~\cite{nweke2018deep}. In contrast, the End-to-End approach incorporates feature learning while adjusting the discriminative classifier.
	
\end{itemize}

If we look at the TSC problem, we see that the CNNs are the most used architectures, mainly due to their robustness and their relatively short training time, compared to other types of networks. One of the best-known architectures is the Residual Networks (ResNets)~\cite{wang2017time}. This proposal adds linear shortcuts for the convolutional layers, potentially improving the accuracy of the model.

\section{Time series complexity measures and features}
\label{CMF_proposal}
The complexity of a time series represents the interrelationship that exists between its different elements. A greater number of interrelations between the elements of a time series indicates a greater complexity. Once these interrelations have been found and understood, we can try to find the mechanisms that produce this complexity. In this way, it is possible to explain the behavior of a time series based on these mechanisms. In other words, these interrelations are characteristic of the time series.

The features of a time series explain certain behavioral characteristics of the time series itself. The features that traditionally have been used in the process of analysis of a time series as seasonality, trend, stationarity, among others, are well documented~\cite{kang2018efficient}\cite{baldan2018Forecasting}. These types of characteristics can describe the behavior of a time series efficiently. There are other types of characteristics that provide small pieces of information about the behavior of a time series, such as mean, maximum value, minimum value, variance, among others. Although the latter is not usually employed in the analysis process, they are features that may be especially useful depending on the problem. For example, in a classification problem where time series of different classes have significant differences in their value ranges, the mean can be very helpful.

This work presents a novel ensemble of complexity measures and features of time series, aimed at solving problems of classification of time series by applying traditional classification algorithms. It also aims to obtain interpretable results. The characteristics selected in this paper, composed of complexity measures and time series features, are based on information theory and seek to provide greater knowledge about the underlying structure of the processed time series.  A set of characteristics, based on measures of complexity, is summarized in Table~\ref{cm}.

\setlength{\tabcolsep}{2pt}
\begin{table}
	\centering
	\caption{Complexity measures selected.}
	\label{cm}	
	\begin{tabular}{llp{6.5cm}c}
		\toprule
		Char. & Name                    & Description                                                                   & Ref.                \\
		\midrule
		$C_{1}$              & lempel\_ziv             & LempelZiv (LZA)                                                               & \cite{lempel1976complexity}            \\
		$C_{2}$              & aproximation\_entropy   & Aproximation Entropy                                                          & \cite{steven1991approximate}       \\
		$C_{3}$              & sample\_entropy         & Sample Entropy (DK Lake in Matlab)                                            & \cite{richman2000physiological}       \\
		$C_{4}$              & permutation\_entropy    & Permutation Entropy (tsExpKit)                                                & \cite{christoph2002permutation}            \\
		$C_{5}$              & shannon\_entropy\_CS    & Chao-Shen Entropy Estimator                                                   & \cite{chao2003nonparametric}                     \\
		$C_{6}$              & shannon\_entropy\_SG    & Schurmann-Grassberger Entropy Estimator                                       & \cite{schurmann1996entropy}                     \\
		$C_{7}$             & spectral\_entropy       & Spectral Entropy                                                              & \cite{zhang2008}                \\
		$C_{8}$             & nforbiden               & Number of forbiden patterns                                             & \cite{amigo2010}                     \\
		$C_{9}$             & kurtosis                & Kurtosis, the "tailedness" of the probability distribution                   & \cite{decarlo1997}                     \\
		$C_{10}$             & skewness                & Skewness, asymmetry of the probability distribution                          & \cite{vcisar2010}                     \\
		\bottomrule
	\end{tabular}
\end{table}

In addition to the features mentioned above, we have added a set of time series features. It has been selected based on its theoretical basis, also taking into account its historical importance in the field of time series and its interpretability~\cite{kang2018efficient}.
This set of measures, based mostly on typical characteristics of time series, is summarized in Table~\ref{cmh}.

\setlength{\tabcolsep}{2pt}
\begin{table*}
	\centering
	\footnotesize
	\caption{Time series features selected.}
	\label{cmh}
	\begin{adjustwidth}{-6em}{-6em}
	\begin{tabular}{lp{3cm}p{11.25cm}}
	\toprule
	Char. & Name                    & Description \\
	\midrule
	$C_{11}$ & x\_acf1                     & First autocorrelation coefficient \\
	$C_{12}$ & x\_acf10                    & Sum of squares of the first 10 autocorrelation coefficients                                                                                                                     \\
	$C_{13}$ & diff1\_acf1                 & Differenced series first autocorrelation coefficients                                                                                                                           \\
	$C_{14}$ & diff1\_acf10                & Differenced series sum of squares of the first 10 autocorrelation coefficients                                                                                                  \\
	$C_{15}$ & diff2\_acf1                 & Twice differenced series first autocorrelation coefficients                                                                                                                     \\
	$C_{16}$ & diff2\_acf10                & Twice differenced series sum of squares of the first 10 autocorrelation coefficients                                                                                            \\
	$C_{17}$ & max\_kl\_shift              & Maximum shift in Kullback-Leibler divergence between two consecutive windows                                                                                                    \\
	$C_{18}$ & time\_kl\_shift             & Instant of time in which the Maximum shift in Kullback-Leibler divergence between two consecutive windows is located                                                            \\
	$C_{19}$ & outlierinclude \_mdrmd       & Calculates the median of the medians of the values,  while adding more outliers                                                                                                 \\
	$C_{20}$ & max\_level\_shift           & Maximum mean shift between two consecutive windows                                                                                                                              \\
	$C_{21}$ & time\_level\_shift          & Instant of time in which the maximum mean shift between two consecutive windows is located                                                                                      \\
	$C_{22}$ & ac\_9                       & Autocorrelation at lag 9                                                                                                                                                        \\
	$C_{23}$ & crossing\_points            & The number of times a time series crosses the median line                                                                                                                       \\
	$C_{24}$ & max\_var\_shift             & Maximum variance shift between two consecutive windows                                                                                                                          \\
	$C_{25}$ & time\_var\_shift            & Instant of time in which the maximum variance shift between two consecutive windows is located                                                                                  \\
	$C_{26}$ & nonlinearity                & Modified statistic from Ter\"asvirta's test                                                                                                \\
	$C_{27}$ & embed2\_incircle            & Proportion of points inside a given circular boundary in a 2-d embedding space                                                                                                  \\
	$C_{28}$ & spreadrandomlocal \_meantaul & Mean of the first zero-crossings of the autocorrelation function in each segment of the 100 time-series segments of length l selected at random from the original time series   \\
	$C_{29}$ & flat\_spots                 & Maximum run length within any single interval obtained from the ten equal-sized intervals of the sample space of a time series                                                  \\
	$C_{30}$ & x\_pacf5                    & Sum of squares of the first 5 partial autocorrelation coefficients                                                                                                              \\
	$C_{31}$ & diff1x\_pacf5               & Differenced series sum of squares of the first 5 partial autocorrelation coefficients                                                                                           \\
	$C_{32}$ & diff2x\_pacf5               & Twice differenced series sum of squares of the first 5 partial autocorrelation coefficients                                                                                     \\
	$C_{33}$ & firstmin\_ac                & Time of first minimum in the autocorrelation function                                                                                                                           \\
	$C_{34}$ & std1st\_der                 & Standard deviation of the first derivative of the time series                                                                                                                   \\
	$C_{35}$ & stability                   & Stability variance of the means                                                                                                                                                 \\
	$C_{36}$ & firstzero\_ac               & First zero crossing of the autocorrelation function                                                                                                                             \\
	$C_{37}$ & trev\_num                   & The numerator of the trev function, a normalized nonlinear autocorrelation, with the time lag set to 1                                                                          \\
	$C_{38}$ & alpha                       & Smoothing parameter for the level-alpha of Holt’s linear trend method                                                                                                           \\
	$C_{39}$ & beta                        & Smoothing parameter for the trend-beta of Holt’s linear trend method                                                                                                            \\
	$C_{40}$ & nperiods                    & Number of seasonal periods (1 for no seasonal data)                                                                                                                             \\
	$C_{41}$ & seasonal\_period            & Seasonal periods (1 for no seasonal data)                                                                                                                                       \\
	$C_{42}$ & trend                       & Strength of trend                                                                                                                                                               \\
	$C_{43}$ & spike                       & Spikiness variance of the leave-one-out variances of the remainder component                                                                                                    \\
	$C_{44}$ & linearity                   & Linearity calculated based on the coefficients of an orthogonal quadratic regression                                                                                            \\
	$C_{45}$ & curvature                   & Curvature calculated based on the coefficients of an orthogonal quadratic regression                                                                                            \\
	$C_{46}$ & e\_acf1                     & First autocorrelation coefficient of the remainder component                                                                                                                    \\
	$C_{47}$ & e\_acf10                    & Sum of the first then squared autocorrelation coefficients                                                                                                                      \\
	$C_{48}$ & walker\_propcross           & Fraction of time series length that walker crosses time series                                                                                                                  \\
	$C_{49}$ & hurst                       & Long-memory coefficient                                                                                                                                                         \\
	$C_{50}$ & unitroot\_kpss              & Statistic for the KPSS unit root test with linear trend and lag one                                                                                                             \\
	$C_{51}$ & histogram\_mode             & Calculates the mode of the data vector using histograms with 10 bins (It is possible to select a different number of bins)                                                      \\
	$C_{52}$ & unitroot\_pp                & Statistic for the PP unit root test with constant trend and lag one                                                                                                             \\
	$C_{53}$ & localsimple\_taures         & First zero crossing of the autocorrelation function of the residuals from a predictor that uses the past trainLength values of the time series to predict its next value        \\
	$C_{54}$ & lumpiness                   & Lumpiness variance of the variance                                                                                                                                              \\
	$C_{55}$ & motiftwo\_entro3            & Entropy of words in the binary alphabet of length 3.  The binary alphabet is obtained as follows: Time-series values above its mean are given 1, and those below the mean are 0 \\
	\bottomrule			                                              
\end{tabular}
\end{adjustwidth}
\end{table*}

The possible interrelation between the different selected operations has also been analyzed, eliminating those that reached high correlation values.

The objective of using such characteristics is to obtain an alternative and interpretable representation of the behavior of a time series. This representation allows us to use traditional classification algorithms and obtain interpretable results. This way, if a classification algorithm is applied that offers an interpretable model, we can explain the classification based on characteristics that describe the behavior of the processed time series. We can obtain information beyond the simple visual behavior of a time series.

The theoretical explanation of each of the measures has not been included in this paper due to space constraints. For the convenience of the reader, they are available online in the web resource~\footnote{Complexity Measures and Features for Times Series classification. \url{http://dicits.ugr.es/papers/CMFTS/}} associated with this work.

Our proposal consists of a set of characteristics that allow us to classify in a better way the time series and to obtain interpretable results. The pseudocode in Algorithm~\ref{Algorithm1} shows how our proposal works.

\begin{algorithm}[]
	\floatname{algorithm}{Algorithm}
	\caption{Main procedure}
	\label{Algorithm1}
	\small
	\begin{algorithmic}[1]
		\Input{
			\textit{$train$}: train dataframe with (Ts\_class, Ts\_values)\\
			\textit{$test$}: test dataframe with (Ts\_class, Ts\_values) \\
			\textit{$models $}: list of models to be processed \\
		}
		\Output{$output\_data$: a triplet that contains the fitted models, the vectors with the predicted labels and the accuracies obtained \\
			$f\_train$: characteristics train dataframe \\
			$f\_test$: characteristics test dataframe  \\
		}
		\State f\_train, f\_test $\gets$ calc\_cmfts((train.Ts\_values, test.Ts\_values), all)
		\For{each value in (f\_train, f\_test)}
		\If {(is.na(value) $\|$ is.nan(value))} value $\gets$ NA \EndIf
		\EndFor
		
		\For {each column in f\_train}
		
		\If {(count.na(column) $\geq$ (length(column)*0.2))}
		\State  f\_train $\gets$ f\_train[ , -column.index]
		\State  f\_test $\gets$ f\_test[ , -column.index]
		\EndIf
		
		\EndFor
		
		\For {each column in f\_train}
		
		\For {each value in (f\_train[ , column.index], f\_test[ , column.index])}
		
		\If {(is.infinite(value))}
		\If {(value $\geq$ 0)}
		\State \multiline{
			value $\gets$ max(f\_train[ , column.index], ignore.inf)
		}
		\Else
		\State \multiline{
			value $\gets$ min(f\_train[ , column.index], ignore.inf)
		}	
		\EndIf
		\EndIf
		
		\EndFor
		\EndFor
		
		\For {each column in (f\_train, f\_test)}
		\State  column $\gets$ impute.Mean(column)
		\EndFor
		
		\State output\_data $\gets$ NULL
		\For {each model in models}
		\State  fit $\gets$ model.train(f\_train, train.Ts\_class)
		\State	pred $\gets$ fit.predict(f\_test)
		\State 	acc $\gets$ accuracy(pred, test.Ts\_class)
		\State  output$\_$data.add(fit, pred, acc)
		\EndFor
		
		\State \textbf{return} (output\_data, f\_train, f\_test)
	\end{algorithmic}
\end{algorithm}

Our proposal begins with an individual and independent processing of each time series (line 1). The selected set of characteristics is calculated for each time series, obtaining a set of results with as many values as features applied to the time series. By processing the whole set of input time series, we calculate a matrix of values with as many columns as applied features and as many rows as processed time series. This matrix is a representation of the input time series, free of any time dependency, based on the parameters obtained when applying the operations mentioned above. As there is no time dependency in the new dataset, it is possible to use any traditional classification algorithm on this new dataset.

Although most of the proposed characteristics are specially designed to be applied over time series, in some cases, these characteristics may not be defined for some specific time series. In these cases, undesirable values are produced, and we must process them. In the first place, we differentiate between the cases in which we obtain infinite values and those we do not. For this reason, the results obtained are filtered, detecting the cases of noninfinite values and transforming to the same value (lines 2-5) for subsequent elimination or imputation. On the training set, we check for each column (operation applied) that the amount of these values is less than 20\% of the total. In other cases, the column is removed from both the training set and the test set (lines 6-11). Infinite values are identified as positive or negative and replaced by the maximum or minimum value of the corresponding column, respectively, ignoring the infinite values in these calculations (lines 12-22). Imputation of missing values based on the mean is then applied to each column (lines 23-25), eliminating any presence of unwanted values in the datasets.

Since one of our objectives is to obtain interpretable results, in the second part of our proposal, we have selected the main classification algorithms based on trees: C5.0, C5.0 with boosting~\cite{quinlan2014c4}, Rpart~\cite{therneau1997package} and Ctree~\cite{hothorn2015ctree}. We have selected this type of algorithms by the interpretability of the generated models. The accuracy of the models obtained on the test set is an objective indication of the quality or fidelity of the representation obtained by the set of selected features. We initialize a variable that contains the results obtained for each one of the processed models (line 26). In the final part, we calculate each selected model, make the corresponding prediction, and calculate the accuracy. Finally, all these results are stored (lines 27-32). Our proposal returns these results together with the training and test sets with the new calculated characteristics (line 33).

\begin{figure*}[htb!]
	\centering
	\includegraphics[width=1\textwidth]{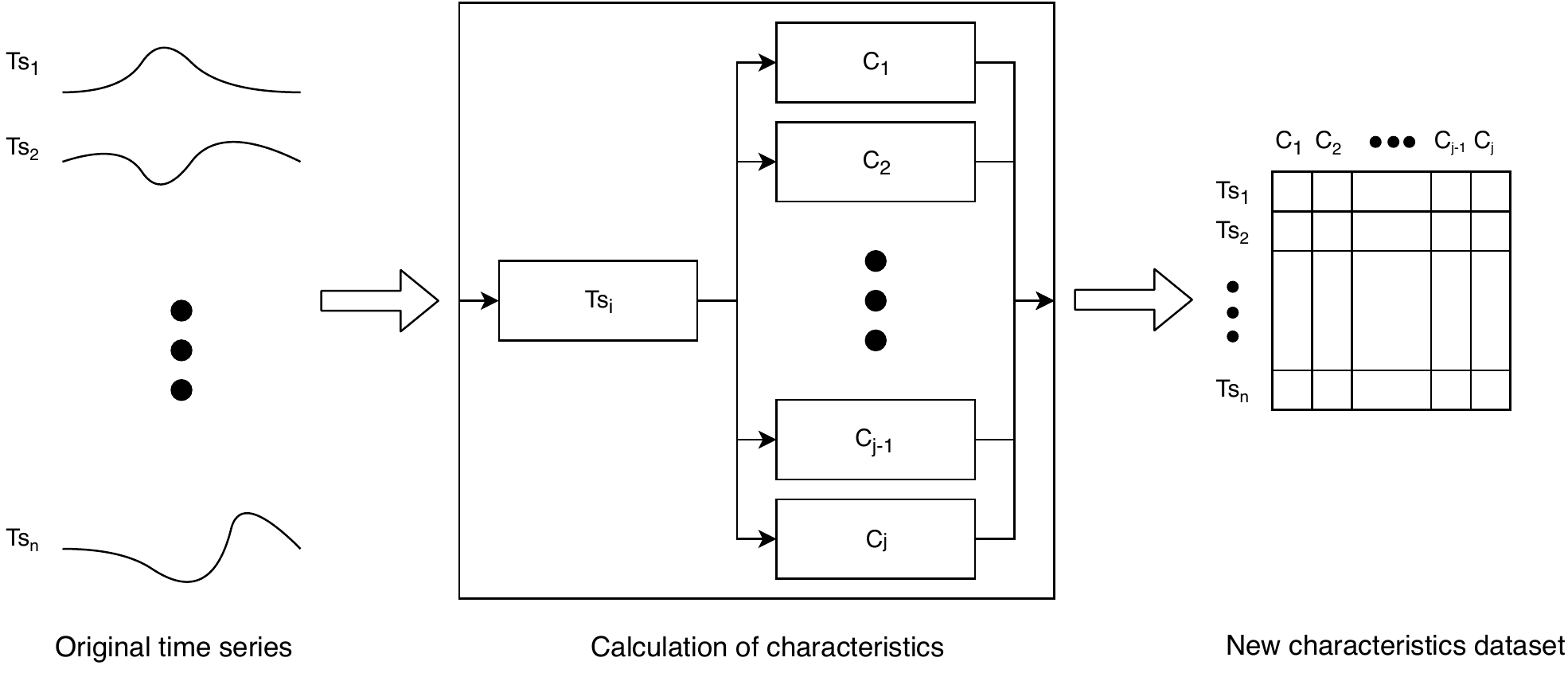}
	\caption{Characteristics calculation workflow.}
	\label{fig:proposaldiagram}
\end{figure*}

Figure~\ref{fig:proposaldiagram} shows, in a graphic form, the process of calculating the characteristics of the time series.

At this point, it is necessary to proceed to the analysis of the trees obtained in search of an interpretable result that, in many cases, is difficult to appreciate in the original time series.

\section{Empirical Study}
\label{Exp}
In this section we evaluate the performance of our proposal. In order to do this, we first show the experimental design carried out followed by the results obtained with their corresponding analysis.

\subsection{Experimental Design}
In this section, we show the measures used to evaluate the performance of our proposal, the datasets processed, the classification models selected and the hardware used in the experimentation.

The source code of our proposal and experimentation has been developed in R 3.4.4 and can be found in the online repository~\footnote{Complexity Measures and Features for Times Series classification. \url{https://github.com/fjbaldan/CMFTS/}}.

\subsubsection{Performance measures}
We have chosen accuracy as a basic measure of performance. Accuracy is calculated as the number of correctly classified instances in the test set divided by the total number of cases in the test set. We use the average rank to compare the performance of the different models against each other. Having over a large number of datasets, very different from each other, a relative performance measure like the rank is one of the best options to make the desired comparison. Since the results can vary greatly from one dataset to another we have chosen to use the Critical Difference diagram (CD)~\cite{demvsar2006statistical}. CD allows making a comparison of results, between the different models, from a statistical point of view. In this diagram, the models linked by a bold line can be considered to have no statistically significant differences in their results at a given confidence level $\alpha$. In this paper, we have chosen a 95\% confidence level setting an $\alpha$ of 0.05. We have used the R \textit{scmamp} package to calculate average rank and the CD. In addition, we include the Win/Loss/Tie ratio to be able to observe in a direct quantitative way the performance of each model in comparison with the rest.

\subsubsection{Datasets}

The used datasets have been extracted from the UCR repository~\cite{dau2019ucr}, which is the reference repository in the field of TSC. It is composed of 128 datasets. The authors of the repository have run the main algorithms of the state of the art of TSC on 112 of the 128 datasets. They eliminated 15 datasets because of containing time series of different lengths and the Fungi dataset because it contains only one instance per class in the training data. Given the great number of algorithms run on these datasets, we can consider the 112 selected datasets as the state of the art in TSC datasets.

\subsubsection{Models}
The main tree classification algorithms have been selected based on their interpretability: C5.0, C5.0 with boosting (C5.0B)~\cite{quinlan2014c4}, Rpart~\cite{therneau1997package}, and Ctree~\cite{hothorn2015ctree}. 1NN+ED, 1NN+DTW(w=100) and 1NN+DTW(w\_learned) applied over the original time series have been included as benchmark methods since they are the benchmark TSC methods. The new representation of time series that we propose in this work offers an additional information about these series that can also be used by less interpretable algorithms to improve the obtained results. For this purpose, classification algorithms with greater complexity and better accuracy performance have been selected like RF~\cite{Breiman2001}, and SVM~\cite{Cortes1995}. We have also added 1NN+ED applied to the proposed features as a benchmark method. We name the models based on the features proposed in this work following the CMFTS+Model pattern, for example, CMFTS+RF, CMFTS+C5.0, etc.

In order to evaluate our proposal, we have selected only the main algorithms of the state of the art that have been run on the 112 datasets previously mentioned. The algorithms selected are: HIVE-COTE, STC, ResNet, WEASEL, BOSS, cBOSS, c-RISE, TSF, and Catch22. We do not include the model FEARS because there are not public results over the 112 selected datasets, and we were not able to reproduce the results of the original work.

\subsubsection{Hardware}
For our experiments, we have used a server with the following characteristics: 4 × Intel(R) Xeon(R) CPU E5-4620 0 @ 2.20GHz processors, 8 cores per processor with HyperThreading, 10 TB HDD, 512 GB RAM. We have used the following software configuration: Ubuntu 18.04, R 3.6.3.

\subsection{Results}
In this section, we show and evaluate the results obtained by our proposal both in terms of performance (Section~\ref{performanceResults}) and interpretability (Section~\ref{Inter}). Since the complete empirical results are too extensive to include in the paper, we have put just a summary. The complete set is available at web resource\footnote{Complexity Measures and Features for Times Series classification. \url{http://dicits.ugr.es/papers/CMFTS/}} associated to this work.

\subsubsection{Performance results}
\label{performanceResults}

Table~\ref{totalAccTree} shows the results obtained for the 112 datasets processed. We show the average accuracy, average rank, and Win/Loss/Tie Ratio, for all the feature-based learning models (CMFTS) proposed in this paper and the benchmark models in TSC.

\begin{table}[htb!]
	\centering
	\caption{Comparative results of the proposed feature-based models (CMFTS) and the TSC benchmark models. The best results are stressed in bold.}
	\label{totalAccTree}
	\begin{tabular}{|l|l|l|l|}
		\toprule
		Model & Average Acc. & Average Rank & W/L/T Ratio \\
		\midrule
		CMFTS+C5.0           & 0.724      & 6.442       & 3/109/2   \\
		CMFTS+C5.0B          & 0.766      & 4.263       & 12/100/3  \\
		CMFTS+Rpart          & 0.682      & 7.071       & 4/108/1   \\
		CMFTS+Ctree          & 0.652      & 7.683       & 4/108/2   \\
		CMFTS+RF             & \textbf{0.807}      & \textbf{2.567}       & \textbf{48/64/4}   \\
		CMFTS+SVM            & 0.764      & 4.21        & 14/98/5   \\
		CMFTS+1NN-ED         & 0.737      & 5.996       & 8/104/4   \\
		1NN-ED               & 0.694      & 6.388       & 9/103/9   \\
		1NN-DTW (learned\_w) & 0.752      & 4.71        & 23/89/11  \\
		1NN-DTW (w=100)      & 0.73       & 5.67        & 16/96/5  \\
		\bottomrule
	\end{tabular}
\end{table}

If we look at the results of the average rank, Table~\ref{totalAccTree}, we see that the CMFTS+RF model obtains the best results, followed by CMFTS+SVM, CMFTS+C5.0B, and 1NN-DTW (learned\_w). This shows that more complex models such as RF, C5.0B, SVM, and 1NN-DTW (learned\_w) offer better results than more simple models such as C5.0, Rpart, and Ctree. This behavior is also visible in the Win/Loss/Tie Ratio, where CMFTS+RF is the best model, with 48 wins, followed by 1NN-DTW (learned\_w) with 23 wins. The third, fourth and fifth places are taken by 1NNN-DTW (w=100) (16 wins), CMFTS+SVM (14 wins), and CMFTS+C5.0B (12 wins), respectively.

\begin{figure}[htb!]
	\centering
	\includegraphics[width=0.8\textwidth]{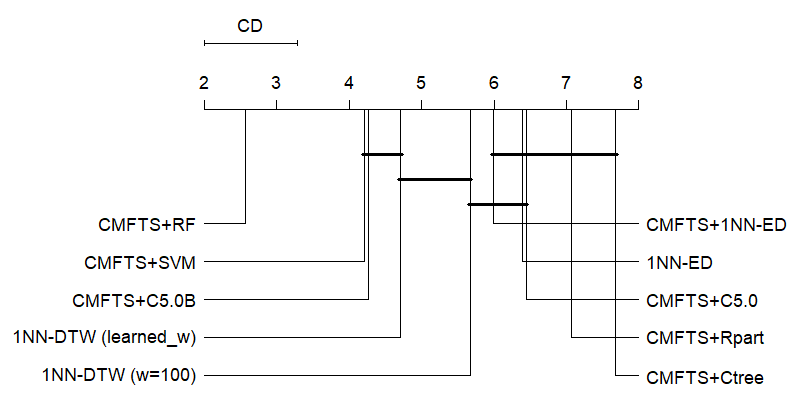}
	\caption{Critical Difference diagram between the proposed feature-based models (CMFTS) and the TSC benchmark models, confidence level of 95\%.}
	\label{fig:CD_tree}
\end{figure}

In order to make a statistically robust comparison between the different models, we used the CD shown in Figure~\ref{fig:CD_tree}, with a confidence level of 95\%. The CD diagram shows that there is no statistical relationship between CMFTS+RF and the other models, being CMFTS+RF the model most interesting of the tested set. We also see how there are no statistically significant differences between the CMFTS+SVM, CMFTS+C5.0B, and 1NN-DTW (learned\_w) models, being the CMFTS+C5.0B model the one with a higher degree of interpretability. Those results allow us to aspire to have interpretable models with competitive results. Finally, we see how CMFTS+1NN-ED slightly improves the results of its direct competitor 1NN-ED and the remaining of the tree-based models (C5.0, Rpart, and Ctree). But the differences are not significant from a statistical point of view.

\begin{figure}[htb!]
	\centering
	
	\subfloat[Full UCR repository, 112 datasets.]{
		\label{mainCD:all}
		\includegraphics[width=0.8\textwidth]{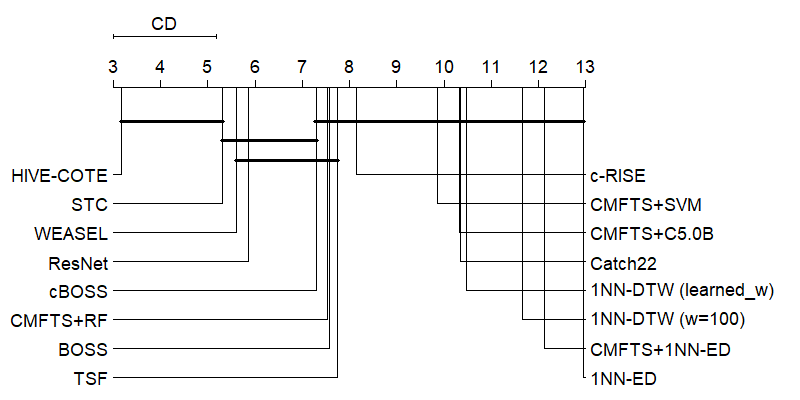}}
	
	\subfloat[Datasets with 100 or more instances, 76 datasets.]{\label{mainCD:100}\includegraphics[width=0.8\textwidth]{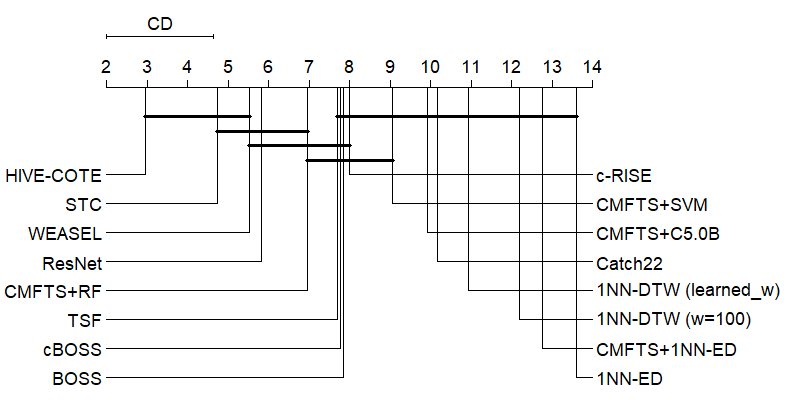}}	
	
	\subfloat[Datasets with 500 or more instances, 25 datasets.]{\label{mainCD:500}\includegraphics[width=0.8\textwidth]{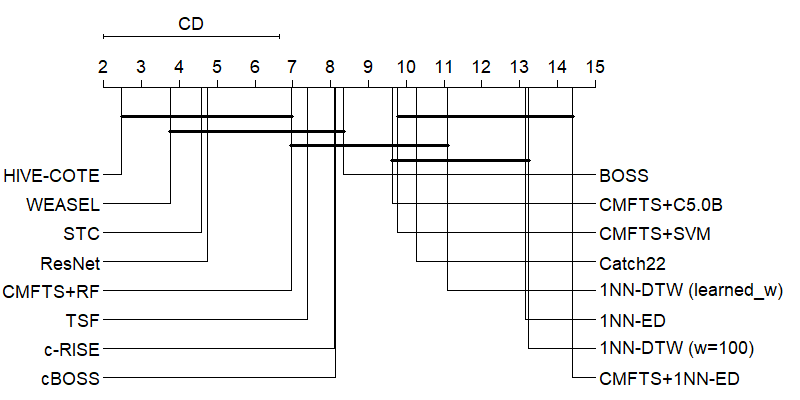}}
	
	\caption{Critical Difference diagrams between the proposed feature-based models (CMFTS) and the TSC state of the art models, confidence level of 95\%. Different scenarios.}
	\label{figCD:main}
\end{figure}

Once the best models of our proposal have been identified, we will compare them with the best models of the state of the art. The best models of our proposal selected for this comparison are CMFTS+RF, CMFTS+SVM, CMFTS+C5.0B, and CMFTS+1NN-ED. CMFTS+RF and CMFTS+SVM are the models that obtain the best results, although their interpretability is reduced. CMFTS+C5.0B is the most interpretable model with the best results if we compare it with the rest of the tree-based models. CMFTS+1NN-ED is a simple model that we can use as a benchmark. As in the previous case, for a first analysis, we use a table with the results of average accuracy, average rank, and Win/Loss/Tie Ratio, Table~\ref{totalAccAll}. In addition, to carry out an analysis from a statistical point of view we use the CD, Figure~\ref{figCD:main}.

In Table~\ref{totalAccAll}, we see the HIVE-COTE algorithm has the best results in average rank, average accuracy, and win/loss/tie ratio. This algorithm should be used whenever possible. STC is the second method with the lowest average rank and higher average accuracy, but the third in the win/loss/tie ratio. STC can obtain good results in a great number of cases, but not the best results. This behavior indicates that STC offers competitive and robust results in different fields. WEASEL has a behavior very similar to STC. It is the third method in the average rank results, and it has a win/loss/tie ratio and average accuracy results lower but very close to the STC results. For the same win/loss ratio values, WEASEL obtains a higher number of ties than STC. Both methods offer a good start point. RestNet is the fourth method in average rank, but the second one on the win/loss/tie ratio. This behavior indicates that it works better in certain cases, obtaining the best results in a higher number of cases in comparison with STC and WEASEL. In another way, RestNet has worse average performance. If we analyze our proposals, we could observe that CMFTS+RF offers the best results on the average rank, win/loss/tie ratio, and average accuracy.

\begin{table}[htb!]
	\centering
	\caption{Comparative results of the proposed feature-based models (CMFTS) and the TSC state of the art models. The best results are stressed in bold.}
	\label{totalAccAll}
	\begin{tabular}{|l|l|l|l|}
		\toprule
		Model & Average Acc. & Average Rank & W/L/T Ratio \\
		\midrule
		CMFTS+RF             & 0.807      & 7.531       & 10/102/5  \\
		CMFTS+SVM            & 0.764      & 9.871       & 5/107/2   \\
		CMFTS+C5.0B          & 0.766      & 10.321      & 1/111/0   \\
		CMFTS+1NN-ED         & 0.737      & 12.116      & 4/108/4   \\
		BOSS                 & 0.815      & 7.58        & 12/100/12 \\
		Catch22              & 0.769      & 10.353      & 3/109/2   \\
		cBOSS                & 0.818      & 7.29        & 15/97/13  \\
		c-RISE               & 0.79       & 8.156       & 7/105/5   \\
		HIVE-COTE            & \textbf{0.864}      & \textbf{3.17}        & \textbf{42/70/17}  \\
		ResNet               & 0.82       & 5.866       & 33/79/9   \\
		STC                  & 0.845      & 5.308       & 18/94/9   \\
		TSF                  & 0.786      & 7.741       & 9/103/7   \\
		WEASEL               & 0.834      & 5.603       & 18/94/12  \\
		1NN-ED               & 0.694      & 12.955      & 1/111/1   \\
		1NN-DTW (learned\_w) & 0.752      & 10.473      & 4/108/3   \\
		1NN-DTW (w=100)      & 0.73       & 11.665      & 8/104/5  \\
		\bottomrule
	\end{tabular}
\end{table}

If we analyze Figure~\ref{figCD:main}, we can observe statistical relationships between our proposal CMFTS+RF and the algorithms HIVE-COTE, STC, and WEASEL, with some conditions. In Figure~\ref{mainCD:all}, there are four principal subgroups of proposals without statistical differences between their results over the 112 selected datasets. In this case, HIVE-COTE and STC compose the group with the best results. We can observe that the last group is composed of twelve proposals, which is an interesting behavior. We see how CMFTS proposals are included in this group, but CMFTS+RF is included in another group where its results do not differ statistically from those obtained by WEASEL. If we increase the minimum number of instances per dataset, the observed subgroups can vary significantly. Normally, the features-based approach performs worse in datasets with a low number of instances. In Figure~\ref{mainCD:100}, using datasets with 100 instances or more, we have five different subgroups of models. Now, the best group is composed of HIVE-COTE, STC, and WEASEL. In this case, we see how the results of our best proposal, CMFTS+RF, have not statistical differences with STC and WEASEL models, which are included in the first group. In Figure~\ref{mainCD:500}, using datasets with 500 instances or more, we see how the results of CMFTS+RF have not statistical differences with the best model, HIVE-COTE, since CMFTS+RF has been included in the first group. In this case, we can see how WEASEL is the second best model. Those results support the idea that the number of instances affects the results of the features-based methods.

\subsubsection{Interpretability}
\label{Inter}

In this section, we analyze the interpretability of the results obtained by our proposals. We also see the advantages of our proposal in terms of the robustness of results.

\begin{figure}[htb!]
	\centering
	
	\subfloat[GunPoint classes example.]{\label{main:a}\includegraphics[width=0.6\textwidth]{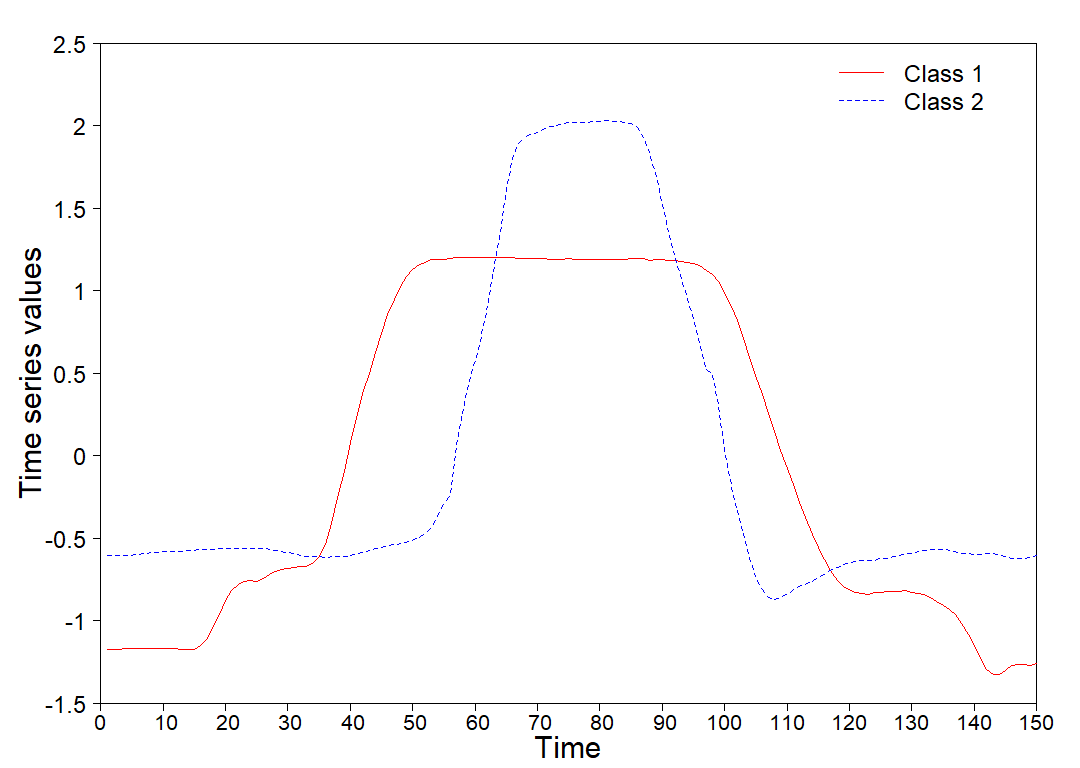}}
	
	\subfloat[GunPoint example, first C5.0B tree with time series measures.]{\label{main:b}\includegraphics[width=0.6\textwidth]{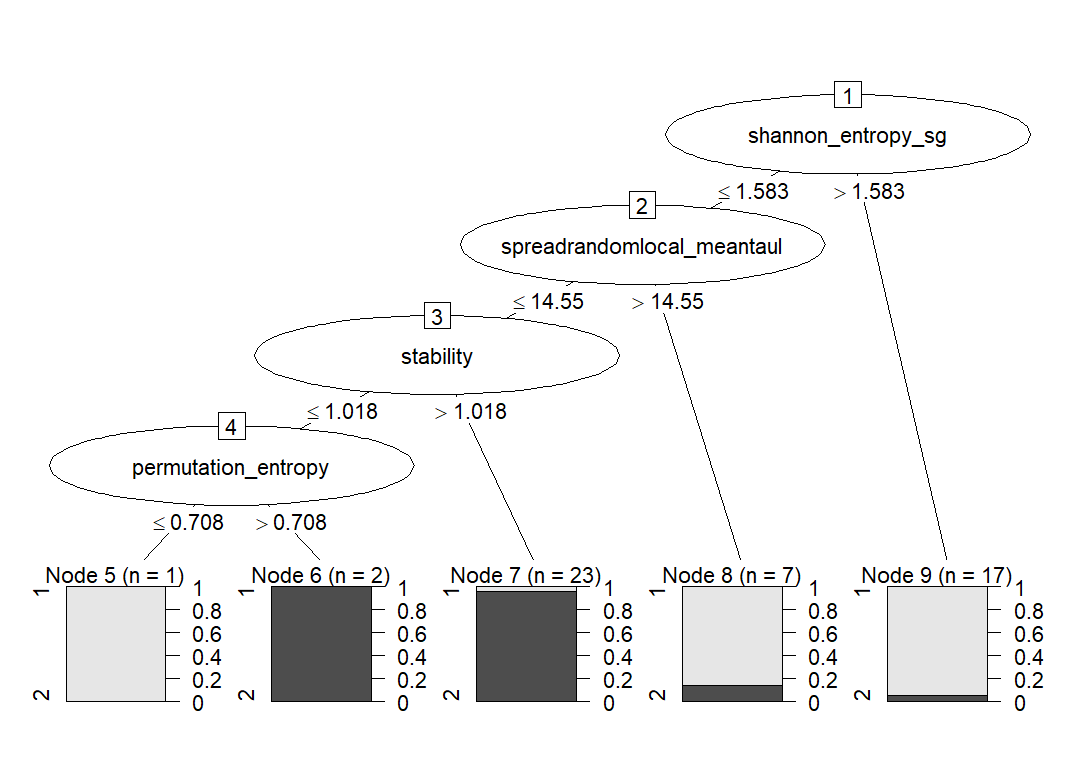}}	
	
	\subfloat[GunPoint example, first C5.0B tree with time series original values.]{\label{main:c}\includegraphics[width=0.6\textwidth]{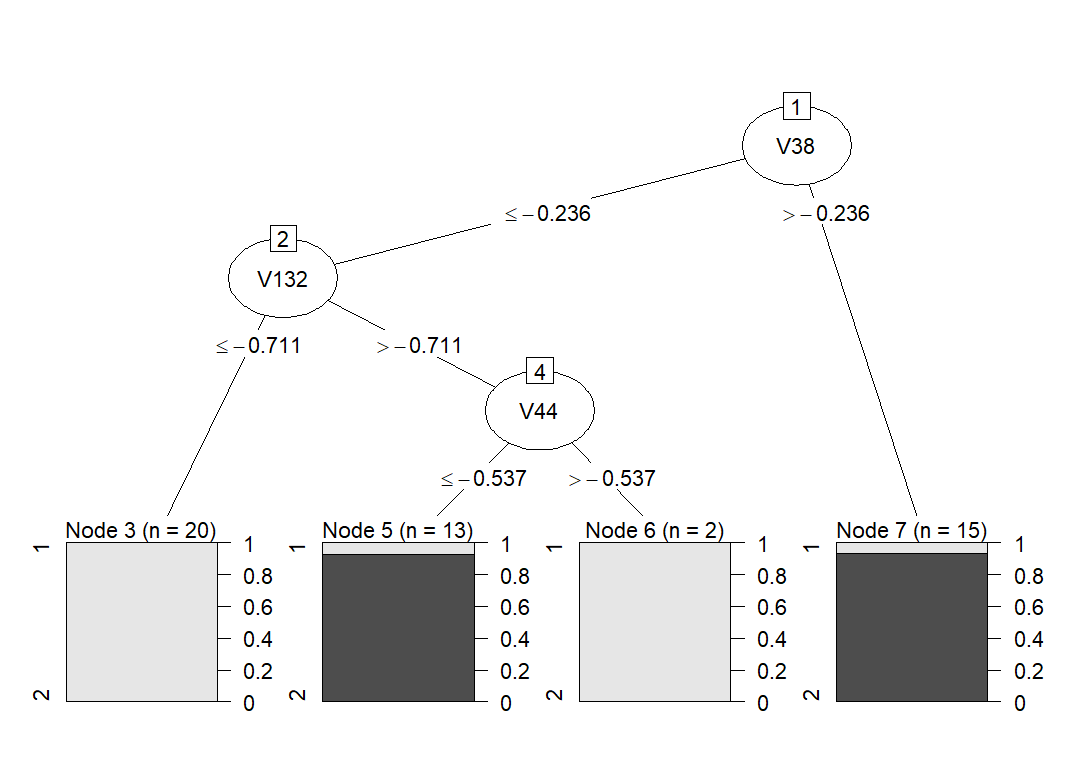}}	
	\caption{Interpretability GunPoint dataset example.}
	\label{fig:main}
\end{figure}

In Figure~\ref{main:a}, we show an example of each of the classes present in the TSC problem called \textit{GunPoint}. It is a problem that differentiates whether a person has a weapon in his hands or not. The time series that compose this problem comes from the center of mass of the right hand of the person holding or not a weapon. Visually it is appreciated that, in the case of having a gun, the peak present in this temporal series is more pronounced than in the case of not having it.

In Figure~\ref{main:b}, we see the first classification tree C5.0 obtained by our proposal, CMFTS+C5B. In this tree, we observe how two features like the \textit{stability}, as the variance of the means obtained from tiled windows, and the \textit{shannon entropy SG}, with Bayesian estimates of the bin frequencies using the Dirichlet-multinomial pseudo-counting model, can differentiate a large part of the cases that belong to a class. If we compare these results with Figure~\ref{main:c}, where the values of some instants of time are the ones that determine if a case belongs to different classes, we can see how our proposal offers a robust behavior to problems as simple as the desynchronization of the temporal series.

The interpretability of the results is strongly linked to the importance given by each algorithm to each of the input features, whenever it is possible. For this reason, we have selected our best proposal, CMFTS+RF, that measures the importance of each feature through the Gini Index~\cite{ceriani2012origins}. We have analyzed the accumulated importance of each feature over the 112 datasets and the importance of each feature in each dataset.

\begin{figure*}[htb!]
	\centering
	\includegraphics[width=1\textwidth]{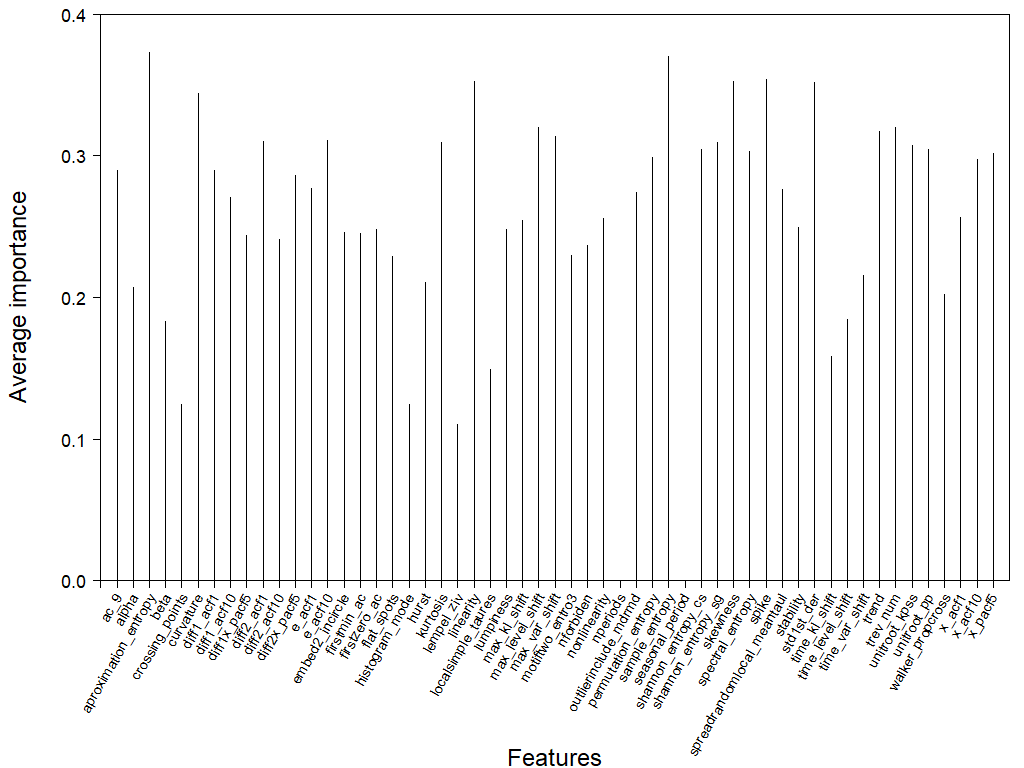}
	\caption{Average importance of features above all datasets.}
	\label{fig:AICA}
\end{figure*}

Figure~\ref{fig:AICA} shows the mean results of the importance of the features obtained on the 112 datasets used. We see how characteristics related to entropy, such as \textit{sample\_entropy} and \textit{aproximation\_entropy} achieve the highest valuation in importance. Interpretable characteristics such as \textit{linearity}, \textit{curvature},  \textit{spike}, and \textit{skewness} would occupy the following positions of importance. On the other hand, we can see two characteristics that have zero importance: \textit{nperiods} and \textit{seasonal\_period}. Given the high number of datasets, a no-preprocessing of the data approach has been chosen, specifying a zero frequency for every time series. This causes the calculation of \textit{nperiods} and \textit{seasonal\_periods} to always get the same value. In a real case, different parameters can be specified that would allow different values to be obtained in these characteristics. The previous characteristics are especially interesting in the field of time series, so we have decided to keep them in the CMFTS package.

\begin{figure*}[htb!]
	\centering
	\includegraphics[width=1\textwidth]{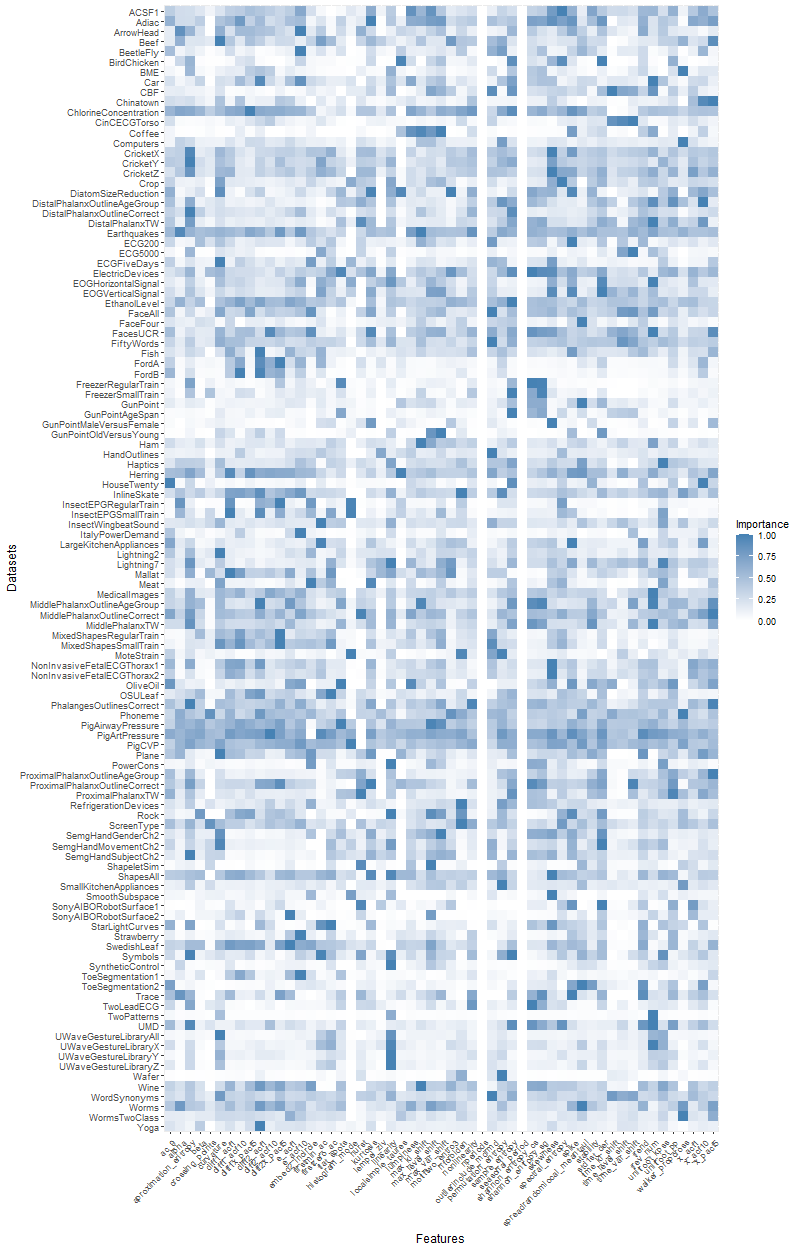}
	\caption{Heat map of the importance of characteristics by dataset.}
	\label{fig:Heatmap}
\end{figure*}

\begin{figure*}[htb!]
	\centering
	\includegraphics[width=1\textwidth]{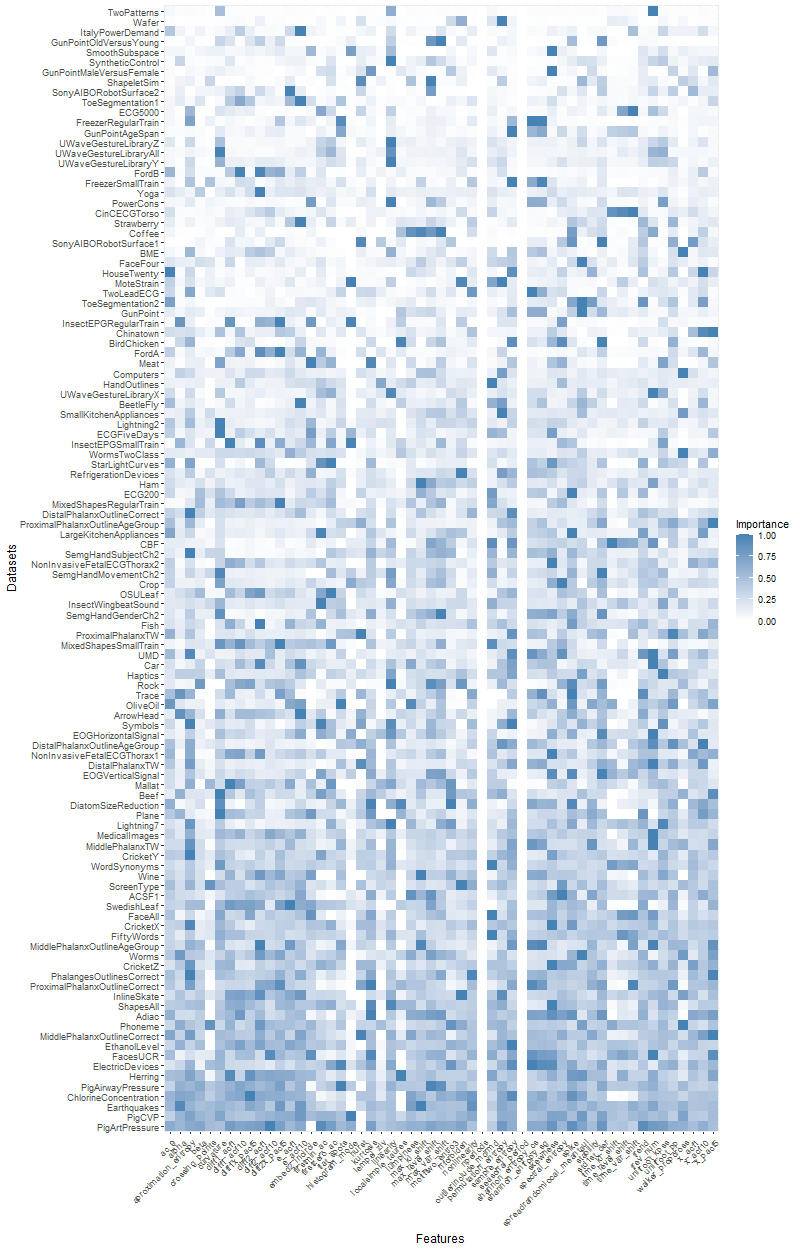}
	\caption{Heat map of the importance of characteristics by dataset, sorted by accumulated importance.}
	\label{fig:HeatmapOrdered}
\end{figure*}

\begin{figure}[htb!]
	\centering
	\includegraphics[width=1\textwidth]{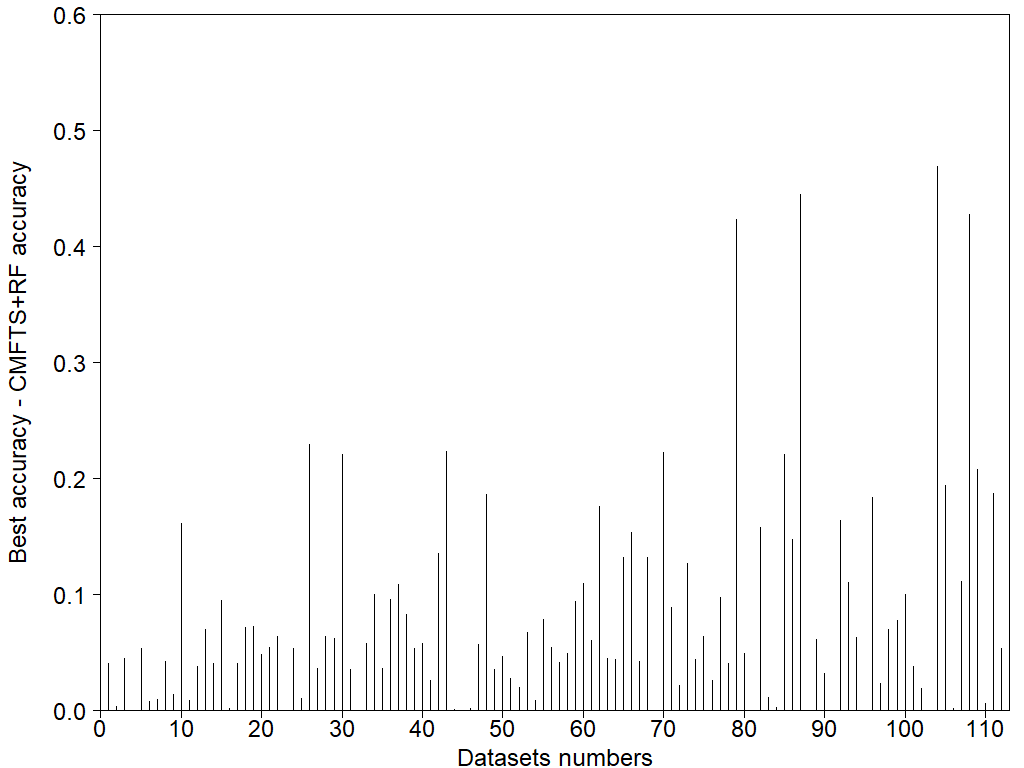}
	\caption{Accuracy differences between CMFTS+RF and the best algorithm on each dataset. The datasets are sorted like Figure~\ref{fig:HeatmapOrdered}.}
	\label{fig:AccDiff}
\end{figure}

We use a heat map to be able to analyze the importance of each feature on each dataset, Figure~\ref{fig:Heatmap}. As we can see in Figure~\ref{fig:Heatmap}, there are a lot of differences in the feature importance scores between different datasets. It means that each problem has very several characteristics and behaviors, so we need different features to extract the right information on each dataset. We can differentiate into two big groups of datasets. The first one which we need a small number of features to obtain the desired information. So, our models can obtain good enough results with this small subset of features, even if these results are not the best. The second one which our model uses a lot of features. In this case, it might be because the problem is very complex, and we need a lot of information to obtain good results. Or the features are not good enough to obtain the needed information to resolve the problem, and the model uses a lot of them trying to obtain good results. If we sort the datasets from Figure~\ref{fig:Heatmap} in an increasing way based on the accumulated importance of the features, we can observe both groups in an easy way, Figure~\ref{fig:HeatmapOrdered}. At the top of the heat map, we can see the datasets in which our model uses a small set of features. At the bottom, we are able to see the datasets in which our proposal needs to use a lot of features. On the datasets in the order of Figure~\ref{fig:HeatmapOrdered}, if we calculate the difference between the best case of each dataset and our best model (CMFTS+RF), Figure~\ref{fig:AccDiff}, we see that this difference is lesser in the datasets at the top of Figure~\ref{fig:HeatmapOrdered}. That means that in the cases in which our model uses a small subset of features, it is able to obtain very close results to the best algorithm. These results reinforce the original idea of this proposal to obtain competitive results with simple and interpretable models.

\section{Conclusion}
\label{Concl}
In this work, we have presented a set of characteristics, composed of measures of complexity and representative features of time series, capable of extracting important information from the time series on which they are applied. The proposed set of features makes it possible to tackle TSC problems with traditional classification algorithms, allowing them to obtain useful and interpretable results.

We have published our proposal software to make it accessible and usable for any practitioner or researcher to use. We have published all the results obtained throughout the work to make it fully reproducible. The functioning of our proposal has been tested on 112 datasets obtained from the UCR repository. We have used tree-based classification algorithms due to their high interpretability, and they have been compared with the state of the art TSC algorithms. The results obtained by our proposal have not statistical differences with the third best algorithm of the state of the art of TSC, with a confidence level of 95\%. If we focus our analysis on datasets with more than 500 time series, our proposal obtains results statistically indistinguishable from those obtained by the best state-of-the-art algorithm. This result reinforces the original idea that feature-based methods require a larger number of time series to perform correctly.

Extracting characteristics of interest from time series that are robust and interpretable provides more understandable and even better classification results in some cases. Our proposal demonstrates a robust behavior against typical TSC problems by extracting descriptive characteristics from the time series rather than working on the original series itself. In this way, additional interpretability is achieved, which is especially useful in some problems.

\section{Acknowledgment}
This research has been partially funded by the following grants: TIN2013-47210-P and TIN2016-81113-R both from the Spanish Ministry of Economy and Competitiveness, and P12-TIC-2958 from Andalusian Regional Government, Spain. Francisco J. Bald\'an holds the FPI grant BES-2017-080137 from the Spanish Ministry of Economy and Competitiveness.

We do warm fully acknowledge insightful comments from Rob. J. Hyndman on previous versions of this paper that without doubt greatly contributed to the enhancement of this paper.

The authors would like to thank Prof. Eamonn Keogh and all the people who have contributed to the UCR TSC archive for their selfless work.

\bibliography{cmfts_Baldan}

%
%
%

\end{document}